# Continuity-driven Synergistic Diffusion with Neural Priors for Ultra-Sparse-View CBCT Reconstruction

*Junlin Wang, Jiancheng Fang, Peng Peng, Shaoyu Wang, Qiegen Liu, Senior Member, IEEE*

***Abstract*—The clinical application of cone-beam computed tomography (CBCT) is constrained by the inherent trade-off between radiation exposure and image quality. Ultra-sparse angular sampling, employed to reduce dose, introduces severe undersampling artifacts and inter-slice inconsistencies, compromising diagnostic reliability. Existing reconstruction methods often struggle to balance angular continuity with spatial detail fidelity. To address these challenges, we propose a Continuity-driven Synergistic Diffusion with Neural priors (CSDN) for ultra-sparse-view CBCT reconstruction. Neural priors are introduced as a structural foundation to encode a continuous three-dimensional attenuation representation, enabling the synthesis of physically consistent dense projections from ultra-sparse measurements. Building upon this neural-prior-based initialization, a synergistic diffusion strategy is developed, consisting of two collaborative refinement paths: a Sinogram Refinement Diffusion (Sino-RD) process that restores angular continuity and a Digital Radiography Refinement Diffusion (DR-RD) process that enforces inter-slice consistency from the projection image perspective. The outputs of the two diffusion paths are adaptively fused by the Dual-Projection Reconstruction Fusion (DPRF) module to achieve coherent volumetric reconstruction. Extensive experiments demonstrate that the proposed CSDN effectively suppresses artifacts and recovers fine textures under ultra-sparse-view conditions, outperforming existing state-of-the-art techniques.**

## I. Introduction

In image-guided radiotherapy (IGRT) and long-term lesion follow-up, CBCT provides critical support for precise localization and dynamic monitoring through its three-dimensional imaging capabilities [1]. Particularly in high-resolution imaging scenarios such as evaluating subsolid pulmonary nodules, CBCT clearly reveals three-dimensional morphology and intricate internal structures—including minute nodules within ground-glass opacities and vascular traction signs—that conventional two-dimensional imaging struggles to capture [2], providing critical evidence for early diagnosis and timely intervention.

However, in long-term, frequent follow-ups, the cumulative radiation dose from multiple CBCT scans has become a bottleneck in clinical application. To mitigate radiation exposure, sparse-view acquisition strategies have been widely investigated [3], in which projection data are collected from a significantly reduced number of angular views [4, 5]. However, ultra-sparse-view sampling fundamentally violates the data completeness conditions required by analytical reconstruction algorithms, rendering CBCT reconstruction a highly ill-posed inverse problem [6]. As a result, reconstructed volumes often exhibit severe streak artifacts, loss of structure details [7] and pronounced inconsistencies between adjacent slices [8], substantially limiting their clinical usability.

Achieving high-quality 3D reconstruction from ultra-sparse-view data is a critical challenge in CBCT, as severe projection deficiency violates the data-completeness requirement for stable reconstruction [9]. Solving this ill-posed problem necessitates the introduction of robust, physically-grounded priors to constrain the solution space [10, 11]. Crucially, a physically plausible 3D volume must satisfy two fundamental continuity constraints: angular continuity in the projection domain [12] and inter-slice continuity along the axial direction [13]. These correspond, respectively, to the sinogram smoothness implied by Radon transform [14] and the structural consistency dictated by cone beam geometry [15].

In recent years, researchers have conducted in-depth explorations along these two dimensions, aiming to reconstruct anatomically consistent, texture-preserving high-quality CBCT images from incomplete data. Sparse-view CT sampling disrupts

This work was supported in part by the National Natural Science Foundation of China under Grant U24A20304, Grant U25A20407 and Grant 62201616. (Junlin Wang and Jiancheng Fang are co-first authors.) (Corresponding authors: Shaoyu Wang and Qiegen Liu.)

J. Wang is with School of Mathematics and Computer Sciences, Nanchang University, Nanchang 330031, China. (e-mail: 7812123133@email.ncu.edu.cn).

J. Fang, P. Peng, S. Wang, and Q. Liu are with School of Information Engineering, Nanchang University, Nanchang 330031, China. (e-mail: d5283377@163.com; 6105123139@email.ncu.edu.cn; {wangshaoyu, liuqiegen}@ncu.edu.cn).

2the angular continuity of projection data, causing sinogram discretization artifacts and severe banding in reconstructed images, which remains a core reconstruction challenge [16, 17]. To address this, sinogram-domain reconstruction has evolved through several paradigms: from generative adversarial networks with explicit continuity losses [18], to interpolation methods utilizing non-local self-similarity and higher-order total variation regularization [19], and more recently to advanced diffusion models. The latter includes direction-adaptive sub-Riemannian [20], physics-constrained bidirectional frequency-aware [21], and universal models handling arbitrary sampling rates [22], significantly advancing sinogram inpainting. Nevertheless, these methods predominantly adhere to a decoupled repair-then-reconstruct paradigm, isolating angular continuity enhancement from the final 3D reconstruction. A more fundamental limitation is the prevalent lack of explicit modeling for periodic boundary conditions along the angular dimension, a deviation from the rigorous mathematical requirements of the Radon transform that may constrain both theoretical validity and reconstruction accuracy under extremely sparse sampling.

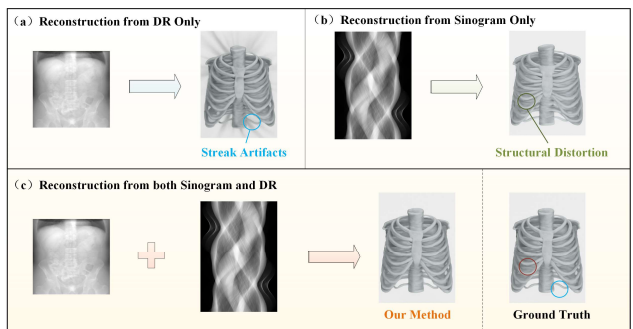

Fig. 1. Comparison of CT reconstruction quality using different projection data sources. (a) Reconstruction from DR-only introduces angular discontinuity, resulting in streak artifacts. (b) Reconstruction from sinogram-only leads to inter-slice discontinuity, causing structural distortion. (c) Our method combines sinogram and DR data for weighted CT reconstruction, effectively eliminating both angular and inter-slice discontinuities, thereby avoiding streak artifacts and structural distortion.

In 3D reconstruction, ensuring anatomical consistency between adjacent slices is crucial for clinical diagnostics. Two primary approaches enforce inter-slice continuity. The first embeds volumetric priors into iterative models, such as hybrid tensor and anisotropic 3D total variation regularizers for spiral CT [23] and 3D low-rank and total variation frameworks for CBCT [24]. These effectively preserve axial consistency but are often tailored to specific scanning geometries, limiting generalizability. The second approach applies post-processing in the image domain, exemplified by deep learning-based slice interpolation [25] and denoising networks with axial consistency discriminators [26]. While enhancing visual coherence, they operate independently of projection data physics, failing to ensure data consistency. This disconnection between model-based methods constrained by geometric specificity and learning-based post-processing lacking physical constraints highlights the need for a novel continuity mechanism that transcends geometric dependencies while remaining coupled to projection data.

While existing sparse-view CBCT reconstruction methods can separately address angular continuity and inter-slice consistency, they struggle to achieve both simultaneously. A visual comparison of reconstructions from single data sources clearly illustrates this dichotomy, as shown in Fig. 1: reconstruction relying solely on the sinogram leads to inter-slice discontinuity and structural distortion, while using only DR projections introduces angular discontinuity and severe streak artifacts. To bridge this gap, this paper proposes a dual-projection reconstruction fusion framework. The main contributions of this work are summarized as follows:

● We propose a continuity-driven synergistic diffusion with neural priors framework (CSDN), which reformulates ultra-sparse-view CBCT reconstruction as a unified continuity modeling problem. CSDN explicitly addresses both angular continuity and inter-slice consistency within a single reconstruction framework.

● A neural prior is introduced to encode a continuous three-dimensional attenuation representation, serving as a structural foundation for the reconstruction process. This enables the synthesis of dense, physically consistent projections from ultra-sparse input data and provides a robust initialization for subsequent refinement.

● We establish a dual-path synergistic refinement diffusion strategy that integrates complementary diffusion processes across projection manifolds. Specifically, a Sinogram Refinement Diffusion (Sino-RD) process that restores angular continuity, and a Digital Radiography Refinement Diffusion (DR-RD) process that enforces inter-slice consistency from the projection image perspective, before being adaptively fused by the Dual-Projection Reconstruction Fusion (DPRF) module. This collaborative refinement enables high-quality, coherent CBCT reconstruction from ultra-sparse-view data.

The remainder of this paper is organized as follows: Section II establishes the necessary background on CBCT and relevant learning-based reconstruction techniques. The architecture and implementation of our Sino-RD and DR-RD methods are detailed in Section III. Section IV validates the framework through comparative experiments and ablation analysis. The paper concludes with a summary in Section V.

## II. RELATED WORK

### A. Overview of Cone-Beam CT

The physical configuration of a CBCT system typically consists of a conical X-ray point source and a two-dimensional detector. Its imaging geometry is defined by several key parameters, including the source-to-origin distance (SOD), source-to-detector distance (SDD), cone angle, and detector pixel size. This geometry determines the system magnification factor, defined as $M_{geo}$=SDD/SOD, which directly correlates with spatial resolution and geometric blurring [27]. Compared to the fan-beam scanning geometries, CBCT is capable of covering the entire imaging volume within a single rotation [28]. This volumetric acquisition characteristic not only enhances scan-ning efficiency but also inherently offers advantages for achieving low dose imaging, particularly when operating under short-arc trajectories [29]. As illustrated in Fig. 2, a typical cone-beam CT configuration is shown.

Under this physical configuration, X-rays transmission through an object follows the Beer-Lambert's law and the measured projection can be modeled as

$$I = I_0 \cdot exp\left(-\int_l \mu(E) \, dl\right), \quad (1)$$

where $I$ represents the transmitted X-ray intensity, $I_0$ is the incident X-ray intensity, $\mu(E)$ denotes the linear attenuation coefficient dependent on energy $E$, and $l$ indicates the ray path from the source to the detector pixel. In practical CBCT systems, the raw projection measurements undergo logarithmic transformation and followed by gain, offset, and other system-specific corrections. These processed data are then used to approximate the line integral required for reconstruction [30].

Despite its practical advantages, CBCT reconstruction is fundamentally challenged by inherent data incompleteness. First, scanning trajectories based on circular or short-arc paths fail to satisfy Tuy's data completeness condition [9], which precludes the derivation of an exact 3D analytical inverse transform. As a result, analytical algorithms such as the FDK can only provide approximate reconstruction results. Second, sparse-view sampling further exacerbates reconstruction artifacts and leads to significant directional resolution deficits [31, 32]. These limitations collectively highlight two central challenges in sparse-view CBCT reconstruction: recovering continuous angular information from sparse sampling, and reconstructing volumetric data that preserve inter-slice consistency.

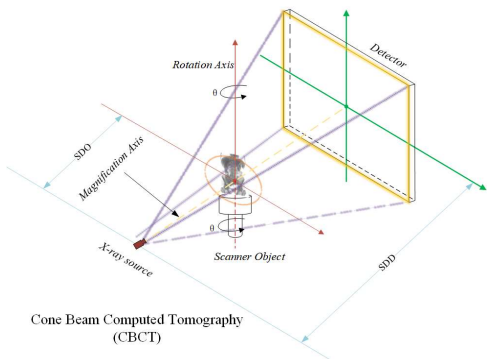

Fig. 2. Schematic of a CBCT system. It includes a conical X-ray source and a flat panel detector, with key parameters such as SOD, SDD and cone angle. CBCT captures volumetric data in a single rotation, offering efficient, low-dose imaging.

### B. Implicit Neural Representations

Ultra-sparse-view CT reconstruction is a severely ill-posed problem, with its core challenge lying in recovering high-fidelity three-dimensional structures from extremely limited projection data [33]. In recent years, implicit neural representations (INRs) have provided new advancements for this field due to their ability to model continuous spatial signals. The success of neural radiance fields (NeRF) demonstrates INRs' exceptional ability to learn continuous 3D scenes from 2D observations [34]. However, their volume-rendering-based mechanism fundamentally diverges from CT's linear physics principles.

To bridge this gap, neural attenuation fields (NAF) [35] were proposed to parameterize the linear attenuation coefficient as a continuous function of spatial coordinates, effectively adapting the INR paradigm to the physics-consistent CT reconstruction framework. Subsequent research has pursued divergent paths to overcome the representational efficiency bottlenecks of the initial NAF framework. Some studies significantly improved efficiency through the introduction of explicit data structures. For instance, NeAT achieved accelerated empty space skipping via explicit octrees [36]. Meanwhile, other efforts were directed toward enhancing the inherent representational capacity and robustness of NAF, with SAX-NeRF setting new benchmarks in novel view synthesis through its innovative architecture and sampling strategy [37]. In parallel, the potential of NAF as a data augmentation tool has emerged. Rather than direct reconstruction, methods such as SNAF leverage novel view synthesis to generate virtual projections that mitigate angular sparsity [38], This paradigm shift transforms INRs from reconstruction engines into prior-informed projection generators.

Despite these advances, most existing INR-based methods remain confined to a single optimization paradigm. While NAF-based direct reconstruction methods achieve high efficiency [36, 37], their tightly integrated representation-physics-optimization frameworks restrict the flexible incorporation of strong prior information, particularly under complex degradation scenarios such as ultra-sparse-view acquisition.

### C. Diffusion Models

Diffusion models learn the latent distribution $p(x)$ through an iterative noise perturbation and denoising process, enabling iterative robust completion of missing information [39]. In the forward diffusion process, perturbates the clean $x_0$ into an approximate noise $x_t$ by progressively introducing Gaussian noise, expressed:

$$q(\mathbf{x}_t|\mathbf{x}_{t-1}) = \mathcal{N}\left(\mathbf{x}_t; \sqrt{1-\beta_t}\mathbf{x}_{t-1}, \beta_t \mathbf{I}\right), \quad (2)$$

where $\mathbf{x}_t$ and $\mathbf{x}_{t-1}$ are the state vectors at diffusion steps $t$ and $t-1$, respectively. $\beta_t$ is the noise variance parameter at step $t$, and $\mathbf{I}$ represents the identity matrix. The backward process progressively denosing through a parameterized network:

$$p_\theta(\mathbf{x}_{t-1}|\mathbf{x}_t) = \mathcal{N}\left(\mathbf{x}_{t-1}; \mu_\theta(\mathbf{x}_t, t), \tilde{\beta}_t \mathbf{I}\right), \quad (3)$$

where $\mu_\theta(\cdot)$ is the parameterized mean function, and $\tilde{\beta}_t$ denotes the noise variance parameter for the reverse diffusion process.

Owing to this generative mechanism, diffusion models are particularly suitable for CT reconstruction tasks involving severely incomplete projection data. Currently, diffusion-based sparse-view CT reconstruction methods primarily follow three technical pathways: sinogram domain, image domain, and dual-domain reconstruction approaches.

Sinogram-domain methods aim to directly complete missing angular information in the projection space. For instance, Guan et al. [40] introduced generative modeling in sinogram domain (GMSD) for sinogram generation, validating this approach's feasibility, Yang et al. [22] proposed a sampling diffusion model (CT-SDM) that reconstructs complete sinograms from arbitrary sparse sampling by simulating sinogram projections.

Image-domain methods perform reconstruction directly in the image space, enhancing visual quality by incorporating real-imaging physics models as constraints alongside diffusion priors. Representative examples include DOLCE, which alternates between physics-model-based data fidelity steps and conditional diffusion prior sampling [41], Han et al. fuse classical primal-dual hybrid gradient (PDHG) optimization with score diffusion models [42].

Dual-domain diffusion models seek to exploit complementary information from both the projection and image domains. Taking dual-domain collaborative diffusion sampling (DCDS) as an example, it employs a bidirectional feedback-based collaborative sampling mechanism: image generation results guide projection completion, while the updated sinogram constrains image reconstruction, forming dual-domain mutual enhancement [43].

In summary, existing methods exhibit limitations in sinogram domain data fidelity, image domain cross-slice consistency, and dual-domain information preservation. Consequently, developing a more robust diffusion framework that can fully exploit original projection information while enforcing physical and structural continuity, which is essential for advancing ultra-sparse-view CBCT reconstruction toward clinically reliable applications.

## III. METHOD

### A. Motivation

CBCT has become an indispensable imaging modality in medical imaging and industrial non-destructive testing due to its high spatial resolution and efficient three-dimensional imaging capability. However, under ultra-sparse-view sampling conditions, the reduction in radiation dose is accompanied by substantial loss of projection data, which impedes the reconstruction of high-quality three-dimensional structures. The core difficulty of ultra-sparse-view CBCT reconstruction lies in constraining the highly ill-posed solution space with physically and structurally meaningful priors. In this work, we argue that a physically plausible CBCT reconstruction should simultaneously satisfy two intrinsic continuity properties imposed by the imaging process: angular continuity in the projection domain and inter-slice continuity in the spatial domain. These two forms of continuity reflect complementary aspects of CBCT imaging physics and geometry, as illustrated in Fig. 3(a) and Fig. 3(b).

From the projection-domain perspective, projection measurements acquired at neighboring view angles are inherently correlated and should vary smoothly according to the Radon transform. Under ultra-sparse angular sampling, large gaps between adjacent views disrupt this smooth evolution, leading to pronounced angular discontinuities in the sinogram and severe streak artifacts in the reconstructed images [44]. Conventional interpolation-based reconstruction methods often struggle to capture the nonlinear variations in projections induced by complex structures, particularly when angular sparsity becomes extreme

On the other hand, the issue of inter-slice continuity stems from the unique 3D cone-beam geometry of CBCT. The general scarcity of high-quality 3D annotated data often forces existing methods to revert to a 2D slice-by-slice reconstruction paradigm. Both end-to-end learning and diffusion model approaches, due to the lack of joint inter-slice modeling, introduce unrealistic axial artifacts and structural discontinuities.

Based on the above analysis, this paper proposes CSDN reconstruction framework. The framework first leverages NAF to construct a continuous 3D scene from ultra-sparse projections, embedding geometric priors. It then employs two core modules for refinement: the Sino-RD, dedicated to enhancing angular continuity, and the DR-RD, dedicated to reinforcing inter-slice consistency, operating in the projection and image domains respectively. Finally, the DPRF module integrates the advantages of angular and inter-slice continuity to produce high-quality images. While the proposed method does not explicitly enforce CBCT data consistency conditions, particularly in cases where orthogonal projections are absent, it meaningfully extends the available sparse data over the complete angular domain. This effectively mitigates the data incompleteness and contributes to a more reliable reconstruction.

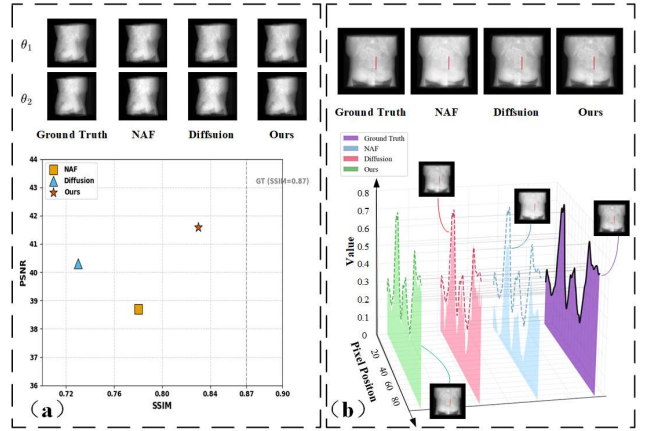

Fig. 3. Motivation schematic. We consider a setting where NAF is trained on full 3D volumes, whereas the diffusion model is trained on axial slices. (a) $\theta_1$ and $\theta_2$ represent two adjacent angles. NAF and diffusion methods may introduce discontinuities between projections, our approach ensures smooth transitions, effectively bridging the gaps in sparse-view data and maintaining structural consistency across projections. (b) NAF projections show similar value accuracy but diverge in distribution, diffusion refinement disrupts data distribution. Our approach maintains both value accuracy and distribution consistency with the ground truth.

### B. Overview of CSDN Reconstruction Framework

This study proposes a dual-projection continuity refinement framework for high-quality 3D CBCT reconstruction from ultra-sparse views, denoted as $N_{sparse}$. The framework, illustrated in Fig. 4(a), is structured around four core components that operate in a coordinated manner to ensure both angular continuity and inter-slice consistency in the final reconstruction.

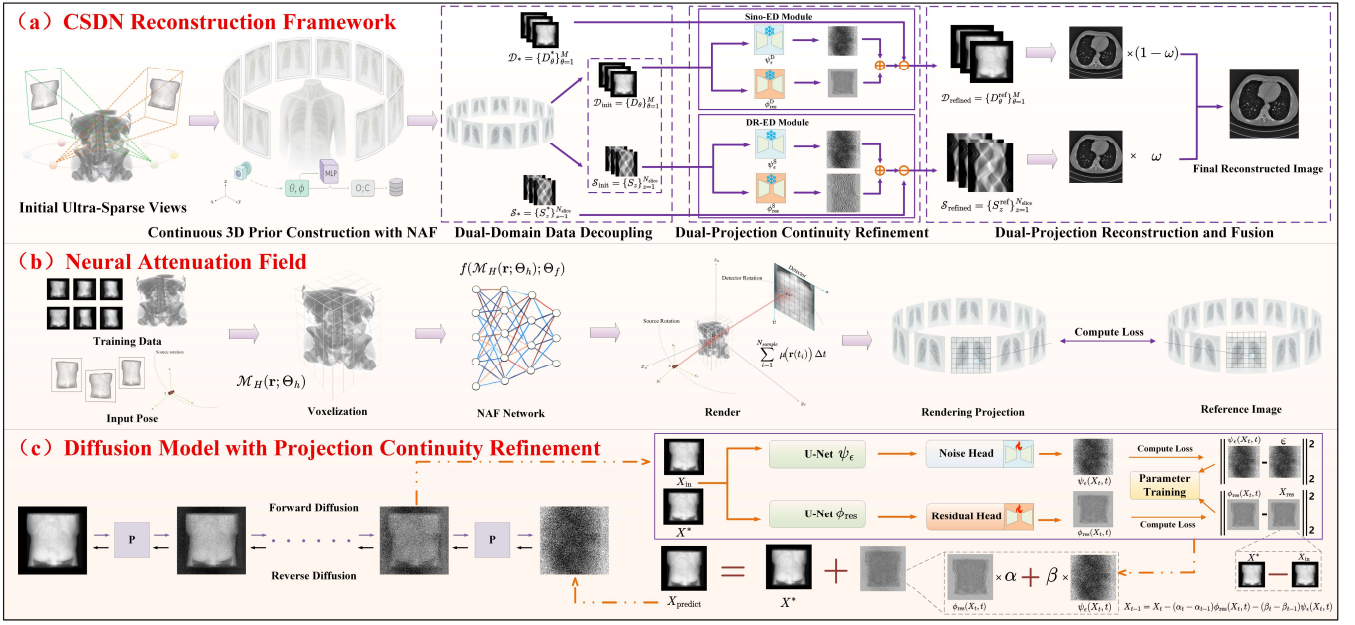

Fig. 4. CSDN Reconstruction Framework. (a) The overall pipeline utilizes ultra-sparse views to train a NAF for constructing a continuous 3D prior. The synthesized dense projections are then processed through two parallel refinement paths: Sino-RD for sinogram-domain angular continuity and DR-RD for DR-domain inter-slice consistency, before being fused into the final reconstruction.(b) The module takes input pose and training data, performs voxelization, and processes them through the NAF network to learn the continuous 3D attenuation representation.(c) This module refines the projection data through a diffusion process, involving forward diffusion to degrade the input and reverse diffusion to recover enhanced, continuous projections.

*1) Continuous 3D Prior Construction with NAF:* The NAF achieves a continuous mapping from the entire 3D spatial coordinates $r = (r_x, r_y, r_z)$ to non-negative linear attenuation coefficients $\mu(r)$ via an implicit neural function $f: \mathbb{R}^3 \rightarrow \mathbb{R}^+$. As a continuous function approximator, this model can effectively capture the piecewise smooth characteristics of real anatomical structures due to its inherent spatial continuity.

To overcome the spectral bias of traditional MLPs in capturing high-frequency details, we integrate learnable multi-resolution hash encoding, adaptively mapping spatial coordinates to high-dimensional features, thus mitigating the network's bias towards low-frequency components. The hash encoder queries and concatenates the coordinates **r** across $L$ scales, producing a feature vector:

$$\mathcal{M}_H(\mathbf{r}; \Theta_h) = [\mathcal{I}(\mathbf{H}_1), \ldots, \mathcal{I}(\mathbf{H}_L)]^T, \quad (4)$$

where $\mathcal{M}_H$ denotes the multi-resolution hash encoding function, $\Theta_h$ represents the parameters of the hash encoder, $\mathbf{H}_i$ denotes the i-th hash table, $\mathcal{I}(\cdot)$ is the hash lookup function, and $L$ indicates the number of encoding scales.

Subsequently, network $f$ maps this high-dimensional feature to the attenuation coefficient of the voxel:

$$\mu(\mathbf{r}) = f(\mathcal{M}_H(\mathbf{r}; \Theta_h); \Theta_f), \quad (5)$$

where $\Theta_f$ denotes the parameters set of the network $f$.

In the cone-beam CT imaging geometry, the X-ray along direction **d** from the source point **o** is parameterized as:

$$\mathbf{r}(t) = \mathbf{o} + t \cdot \mathbf{d}. \quad (6)$$

Its projection value on the detector corresponds to the line integral of the attenuation along the ray path. We uniformly sample $N_{\text{sample}}$ points $\{\mathbf{r}(t_i)\}_{i=1}^{N_{\text{sample}}}$ within the effective range $[t_{\text{near}}, t_{\text{far}}]$, approximating the numerical integration as:

$$P_{\text{pred}} = \sum_{i=1}^{N_{\text{sample}}} \mu(\mathbf{r}(t_i)) \Delta t, \quad \Delta t = \frac{t_{\text{far}} - t_{\text{near}}}{N_{\text{sample}}}, \quad (7)$$

where $t_{\text{near}}$ and $t_{\text{far}}$ are the near and far boundaries of the ray-object intersection, $N_{\text{sample}}$ is the number of sampling points along the ray, and $\Delta t$ represents the sampling interval.

The training objective is to minimize the mean squared error between the predicted projection and the true measurements across the set of rays $\mathcal{R}$:

$$\mathcal{L}_{\text{NAF}} = \frac{1}{|\mathcal{R}|} \sum_{\mathbf{r} \in \mathcal{R}} \| P_{\text{gt}}(\mathbf{r}) - P_{\text{pred}}(\mathbf{r}) \|_2^2, \quad (8)$$

where $\mathcal{R}$ denotes the set of all rays, $|\mathcal{R}|$ represents the total number of rays, and $P_{\text{gt}}(\mathbf{r})$ is the ground-truth projection value for ray **r**.

Through this optimization process, NAF converges to a solution $f_*$ that is guided by both the data fidelity term and the inherent spatial smoothness prior, under sparse observation constraints, as illustrated in Fig. 4(b).

*2) Dual-Domain Data Decoupling:* Based on the fully trained NAF model $f_*$, we synthesize dense projection data covering $M$ angles the entire scanning angle, decoupling the 3D reconstruction task into two independent 2D image enhancement problems, thus optimizing angular continuity and inter-slice continuity separately.

Specifically, given the target angle sequence $\{\theta_k\}_{k=1}^{M}$, where $M$ is much larger than the number of sparse samples and $\theta_k$ denotes the k-th projection angle, our goal is to reconstruct the image using a subset of these angles. For a given $\theta_k$, the projection data at a specific pixel $(u, v)$ is given by the line integral of the attenuation coefficient $\mu$ along the ray $\mathbf{r}(t)$ from $t_{\text{near}}$ to $t_{\text{far}}$. The expression for the projection at this pixel is:

$$P_{\theta_k}(u,v) = \int_{t_{\text{near}}}^{t_{\text{far}}} \mu(\mathbf{r}(t)) \, dt. \quad (9)$$

Then, the collection of all such projection data at each pixel across all angles $\theta_k$ forms the complete projection image $P_{\theta_k}$. By aggregating the projection data from all angles, we obtain the dense projection set $\mathcal{P}_{\text{synth}} = \{P_{\theta_k}\}_{k=1}^M$, which serves as the target projection dataset for the reconstruction process, and the set is fed into two independent data pathways:

**Sinogram Domain Pathway**: The projection data is reorganized into a set of sinograms distributed along the axial position $z$:

$$\mathcal{S}_{\text{init}} = \{S_z \in \mathbb{R}^{M \times A_r}\}_{z=1}^{N_{\text{slice}}}, \quad (10)$$

where $S_z$ represents the sinogram at axial position $z$, $A_r$ denotes the amount of detector rows, and $N_{\text{slice}}$ is the total number of slices.

This structure explicitly constructs a continuous representation along the angle dimension, providing a basis for enhancing angular continuity.

**DR Domain Pathway**: The original organization of the projection data is retained:

$$\mathcal{D}_{\text{init}} = \{D_{\theta_k} \in \mathbb{R}^{A_r \times A_c}\}_{k=1}^M, \quad (11)$$

where $D_{\theta_k}$ represents the DR image at projection angle $\theta_k$, $A_c$ denotes the amount of detector columns.

This pathway aims to preserve the internal structure of the projection images, with its enhancement directly influencing the inter-layer details of the reconstructed volume via the back-projection process.

By decoupling the data in this dual-pathway strategy, we inherit the robust geometric 3D prior provided by the NAF model while simplifying the complex continuity issues into more tractable subproblems in their respective expressive domains.

*3) Dual-Projection Continuity Refinement:* The initial data generated by NAF is processed through our Sino-RD and DR-RD modules: Sino-RD refines angular continuity in the sinogram domain, while DR-RD refines inter-slice continuity in the DR domain. The two pathways adopt independent model parameters and diffusion scheduling coefficients $\alpha_t^S$, $\beta_t^S$ and $\alpha_t^D$, $\beta_t^D$.

**Sino-RD:** This pathway utilizes a sinogram-specific diffusion model $G_S$ to refine the initial sinogram $S_{\text{init}}$. Its core objective is to learn the residual $R_{\text{res}}^S = S^* - S_{\text{init}}$ between the target sinogram $S^*$ and the initial sinogram $S_{\text{init}}$, and to learn the distribution of sinogram noise $\epsilon^S$. This pathway follows its residual diffusion module, with its deterministic forward process defined as:

$$S_t = S_{\text{init}} + \alpha_t^S R_{\text{res}}^S + \beta_t^S \epsilon^S, \quad \epsilon^S \sim \mathcal{N}(0, \mathbf{I}) \quad (12)$$

where $\alpha_t^S$ and $\beta_t^S$ are scheduling coefficients specific to the sinogram-domain path.

During training, the residual network $\phi_{\text{res}}^S$ is optimized to predict the true residual $R_{\text{res}}^S$, and the noise network $\psi_\epsilon^S$ is optimized to predict the true noise $\epsilon^S$. The final refined sinogram is obtained via $S_z^{\text{ref}} = S_{\text{init}} + \phi_{\text{res}}^S(S_0, 0) + \epsilon^S(S_0, 0)$. Consequently, the set of all refined sinograms is $\mathcal{S}_{\text{refined}} = \{S_z^{\text{ref}}\}_{z=1}^{N_{\text{slice}}}$. This process is specifically designed to restore high-frequency nonlinear variations between projection angles, effectively suppressing streak artifacts caused by ultra-sparse sampling.

**DR-RD:** This pathway leverages an independently trained diffusion model $G_D$ on DR images to refine the initial DR image $D_{\text{init}}$. It works by learning to predict the residual $R_{\text{res}}^D = D^* - D_{\text{init}}$ and learning the DR image noise $\epsilon^D$ distribution. Its forward diffusion process employs DR-domain-specific scheduling coefficients:

$$D_t = D_{\text{init}} + \alpha_t^D R_{\text{res}}^D + \beta_t^D \epsilon^D, \quad \epsilon^D \sim \mathcal{N}(0, \mathbf{I}) \quad (13)$$

where $\alpha_t^D$ and $\beta_t^D$ are independent of $\alpha_t^S$ and $\beta_t^S$.

Through its dedicated residual prediction network $\phi_{\text{res}}^D$ and noise prediction network $\psi_\epsilon^D$, this pathway focuses on restoring high-frequency spatial details within DR images, thereby ensuring axial physical consistency in cone-beam geometry. The refined DR image for a single angle is given by $D_\theta^{\text{ref}} = D_{\text{init}} + \phi_{\text{res}}^D(D_0, 0) + \epsilon^D(D_0, 0)$. Thus, the set of refined DR images is $\mathcal{D}_{\text{refined}} = \{D_\theta^{\text{ref}}\}_{\theta=1}^M$.

*4) Dual-Domain Reconstruction and Fusion:* Based on the aforementioned dual-domain projection refinement process, we perform independent FDK reconstructions on the refined sinograms $\mathcal{S}_{refined}$ and DR images $\mathcal{D}_{refined}$. The FDK algorithm converts projection data into image domain volumes through a filtered back-projection process, whose core mathematical expression for a point $\boldsymbol{p}^* = (x^*, y^*, z^*)$ is:

$$V(\mathbf{p}^*) = \int_0^{2\pi} \frac{R^2}{(R+SDD)^2} P(\theta, \gamma(\mathbf{p}^*, \theta)) * h(\gamma) \, d\theta, \quad (14)$$

where $R$ denotes the source-to-rotation-center distance, $\gamma$ is the detector channel angle, $P(\cdot)$ is the preprocessed projection data, $h(\cdot)$ is the ramp filter kernel, and $*$ denotes the convolution operation.

This process yields two optimized reconstructed volumes $V_{\text{sin}}$ and $V_{\text{dr}}$. The volume $V_{\text{sin}}$ benefits from the enhanced angular continuity in the sinogram domain, exhibiting excellent smoothness in the circumferential direction and effectively suppressing stripe artifacts. Meanwhile, $V_{\text{dr}}$ leverages the enhanced inter-slice coherence in the DR domain, maintaining better detail retention across axial slices.

To synergistically utilize the advantages of both domains, we propose a weighted fusion strategy in the image domain. This strategy integrates $V_{\text{sin}}$ and $V_{\text{dr}}$ through a linear combination with weight $\omega$, generating the final fused volume $V_{\text{final}}$:

$$V_{\text{final}}(\mathbf{p}^*) = \omega \cdot V_{\text{sin}}(\mathbf{p}^*) + (1-\omega) \cdot V_{\text{dr}}(\mathbf{p}^*), \quad (15)$$

where $V_{\text{sin}}$ and $V_{\text{dr}}$ are the volumes reconstructed from the refined sinogram and DR domains, respectively, and $\omega \in [0,1]$ is a predefined weighting coefficient hyperparameter that balances angular continuity and inter-slice consistency.

*C. Diffusion Module in Dual-Projection*

The proposed Sino-RD and DR-RD modules are built upon residual diffusion models. The core idea is to train a network to simultaneously predict the residual $I_{res}$ between the target high-quality data $X^*$ and the initial smooth data $X_{init}$ generated by NAF, and the random noise $\epsilon$ introduced during the diffusion process. Here, $I_{res} = X^* - X_{init}$ captures the high-frequency details necessary for reconstruction.

During training, a deterministic forward diffusion process is employed to construct training samples. Specifically, for the true residual $I_{res}$ and random noise $\epsilon \sim \mathcal{N}(0, \mathbf{I})$, the state $X_t$ at step $t$ is formed via linear combination:
$$X_t = X_{init} + \alpha_t I_{res} + \beta_t \epsilon, \qquad (16)$$
where $X_t$ represents the state at diffusion step $t$, $X_{init}$ is the initial data generated by NAF, $I_{res} = X^* - X_{init}$ denotes the target residual, $\alpha_t$ and $\beta_t$ are the scheduling coefficients, and $\epsilon \sim \mathcal{N}(0, \mathbf{I})$ is Gaussian noise.

The training objective is to jointly optimize the parameters of the residual prediction network $\phi_{res}$ and the noise prediction network $\psi_\epsilon$. The update formula for each reverse step from $t$ to $t-1$ is:
$$\mathcal{L}_{res} = \mathbb{E}_{(X_{init}, X^*), t} \| I_{res} - \phi_{res}(X_t, t) \|^2,$$
$$\mathcal{L}_{noise} = \mathbb{E}_{(X_{init}, X^*), \epsilon \sim \mathcal{N}(0, \mathbf{I}), t} \| \epsilon - \psi_\epsilon(X_t, t) \|^2, \qquad (17)$$
where $\psi_\epsilon$ is the noise prediction network, and the expectation is taken over the initial-target pairs $(X_{init}, X^*)$, noise samples $\epsilon$, and diffusion steps $t$.

During the inference stage, a reverse generative process is executed. Starting from an initial state $X_T = X_{init} + \epsilon$ approximating pure noise, iterative denoising of the data is performed using the trained residual and noise prediction networks, as visualized in the refinement process of Fig. 4(c). The formula for each reverse step is:
$$X_{t-1} = X_t - (\alpha_t - \alpha_{t-1})\phi_{res}(X_t, t) - (\beta_t - \beta_{t-1})\psi_\epsilon(X_t, t), \qquad (18)$$
where $\phi_{res}$ and $\psi_\epsilon$ are the residual and noise prediction networks, respectively, and the differences $\alpha_t - \alpha_{t-1}$ and $\beta_t - \beta_{t-1}$ control the step sizes for residual and noise removal.

After $T$ iterations, the final predicted residual $\phi_{res}(X_0, 0)$ and $\psi_\epsilon(X_0, 0)$ are obtained, and the enhanced reconstruction result is computed as follows:
$$X_{ref} = X_{init} + \phi_{res}(X_0, 0) + \epsilon(X_0, 0), \qquad (19)$$
where $X_{ref}$ is the final refined output.

---

**Algorithm 1:** CSDN Reconstruction Framework

**Input:** Ultra-sparse projection set $\mathcal{P}_{sparse} = \{(\theta_i, P_{\theta_i})\}_{i=1}^{N_{sparse}}$, pre-trained NAF $f_*$, pre-trained Sino-RD $G_S$, pre-trained DR-RD $G_D$, fusion weight $\omega$
**Output:** Reconstructed volume $V_{final}$

**Stage 1: Continuous 3D Prior Construction via NAF**
1. **for** each target view angle $\theta_k \in \{\theta_1, \ldots, \theta_M\}$ **do**
2.    Generate per-pixel rays: $\mathbf{r}_{\theta_k}(t) = \mathbf{o}_{\theta_k} + t \cdot \mathbf{d}_{\theta_k}$
3.    Sample along each ray: $\{\mathbf{r}_{\theta_k}(t_i)\}_{i=1}^{N_{sample}}$ with $t_i \in [t_{near}, t_{far}]$
4.    Predict attenuation: $\mu(\mathbf{r}_{\theta_k}(t_i)) \leftarrow f_*(\mathcal{M}_H(\mathbf{r}_{\theta_k}(t_i)))$
5.    Ray integral: $P_{\theta_k} \leftarrow \sum_{i=1}^{N} \mu(\mathbf{r}_{\theta_k}(t_i)) \Delta t_i$
5. **end for**
6. Obtain synthesized dense projections: $\mathcal{P}_{synth} \leftarrow \{P_{\theta_k}\}_{k=1}^{M}$

**Stage 2: Dual-Domain Data Decoupling**
1. Reformulate into sinogram domain: $\mathcal{S}_{init} \leftarrow \{\mathcal{S}_z \in \mathbb{R}^{M \times A_r}\}_{z=1}^{N_{slice}}$
2. Reformulate into DR domain: $\mathcal{D}_{init} \leftarrow \{\mathcal{D}_{\theta_k} \in \mathbb{R}^{A_r \times A_c}\}_{k=1}^{M}$

**Stage 3: Dual-Path Continuity Enhancement**
1. **for** each sinogram $\mathcal{S} \in \mathcal{S}_{init}$ **do**
2.    Predict residual via Sino-RD: $R_{res}^S, \epsilon^S \leftarrow G_S(S_z)$
3.    Enhance angular continuity: $S^{ref} \leftarrow S_z + R_{res}^S + \epsilon^S$
4. **end for**
5. $\mathcal{S}_{refined} \leftarrow \{S_z^{ref}\}_{z=1}^{N_{slice}}$
6. **for** each DR image $\mathcal{D} \in \mathcal{D}_{init}$ **do**
7.    Predict residual via DR-RD: $I_{res}^D, \epsilon^D \leftarrow G_D(D_{\theta_k})$
8.    Enhance inter-slice consistency: $D^{ref} \leftarrow D_{\theta_k} + R_{res}^D + \epsilon^D$
9. $\mathcal{D}_{refined} \leftarrow \{D_\theta^{ref}\}_{\theta=1}^{M}$
10. **end for**

**Stage 4: Dual-Projection Reconstruction and Fusion**
1. Reconstruct sinogram-path volume: $V_{sin} \leftarrow FDK(\{\mathcal{S}_{refined}\})$
2. Reconstruct DR-path volume: $V_{dr} \leftarrow FDK(\{\mathcal{D}_{refined}\})$
3. Fuse volumes: $V_{final} \leftarrow \omega \cdot V_{sin} + (1-\omega) \cdot V_{dr}$
4. **return** $V_{final}$

---

## IV. EXPERIMENTS

### A. Datasets

*a) Public Datasets:* This study primarily employed five publicly available datasets to develop and benchmark the proposed method. They comprise a simulated chest CT volume dataset [45] from the Mayo Clinic's AAPM Low-Dose CT Grand Challenge, and four volumetric datasets (Box, Foot, Head, Jaw) from the Open Science Visualization Dataset [46]. Each simulated 3D volume has dimensions of 256 × 256 × 256 with an isotropic voxel size of 1.0 mm. DR projections were generated using a flat-panel detector of 512 × 512 pixels and a pixel spacing of 1.0 mm, with the SOD and SDD set to 1000 mm and 1500 mm, respectively.

For the AAPM data, ten volumetric slices were available. Among these, eight were used to generate sinogram and DR projections for training the diffusion models in their respective pathways, while the remaining two were reserved for evaluation.

As for the Jaw, Head, Box, and Foot datasets, each provides only a single volumetric slice. Owing to this limited data availability, the same 3D volume was utilized for both training the diffusion model and for weighted reconstruction. To ensure validity, the sets of viewing angles employed during training and inference were mutually exclusive. That is, the sinogram and DR projections used for training and those optimized during weighted reconstruction were acquired from non-overlapping viewpoints.

*b) In-House Experimental Dataset:* In addition to the public benchmarks, we further validated our approach using experimental projection data acquired with our in-house multi-source stationary CBCT system to demonstrate its practical efficacy. As illustrated in Fig. 7, a preserved specimen of Lucanidae, mounted in a glass test tube, was scanned. The imaging geometry and acquisition parameters were as follows: SOD = 103 mm, SDD = 333 mm, and a vertical offset of 1.15 mm along the detector axis. A flat panel detector with 2882 × 2340 pixels and a pixel pitch of 0.0498 mm was used. The X-ray tube was operated at 60 kV and 60 µA. From a full 360° rotation containing 1440 projections, a subset of 50 uniformly spaced views was selected to simulate a sparse-view acquisition scenario. Reconstruction was performed onto a 500 × 500 × 800 voxel grid with an isotropic voxel size of 0.05 mm.

Table I
QUANTITATIVE CT RECONSTRUCTION RESULTS UNDER THE 50-VIEW SETTING. ENTRIES ARE PSNR/SSIM ON SIX DATASETS.

| Method/Dataset | L067 | L096 | Box | Foot | Head | Jaw |
|---|---|---|---|---|---|---|
| FDK [32] | 22.15/0.7015 | 22.91/0.7208 | 26.10/0.7841 | 24.35/0.7133 | 26.97/0.8107 | 24.19/0.6821 |
| SART [47] | 31.42/0.9058 | 31.39/0.9097 | 33.11/0.9471 | 30.22/0.9163 | 34.75/0.9535 | 30.34/0.8944 |
| InTomo [48] | 32.53/0.9387 | 33.42/0.9434 | 33.65/0.9564 | 30.86/0.9218 | 35.73/0.9688 | 32.55/0.9131 |
| NeRF [34] | 32.79/0.9422 | 34.78/0.9579 | 35.40/0.9723 | 31.18/0.9269 | 36.81/0.9777 | 34.12/0.9368 |
| NeAT [36] | 32.74/0.9421 | 33.84/0.9493 | 34.86/0.9684 | 31.19/0.9272 | 35.98/0.9713 | 33.70/0.9308 |
| NAF [35] | 34.91/0.9576 | 35.41/0.9617 | 36.59/0.9764 | 31.70/0.9331 | 37.83/0.9817 | 34.40/0.9387 |
| SAX-NeRF [37] | 35.15/0.9626 | 37.79/0.9770 | 38.17/0.9853 | 32.88/0.9431 | 39.40/0.9881 | 36.15/0.9565 |
| **Ours** | **37.66/0.9774** | **39.23/0.9788** | **41.55/0.9902** | **34.07/0.9565** | **40.51/0.9836** | **39.45/0.9797** |

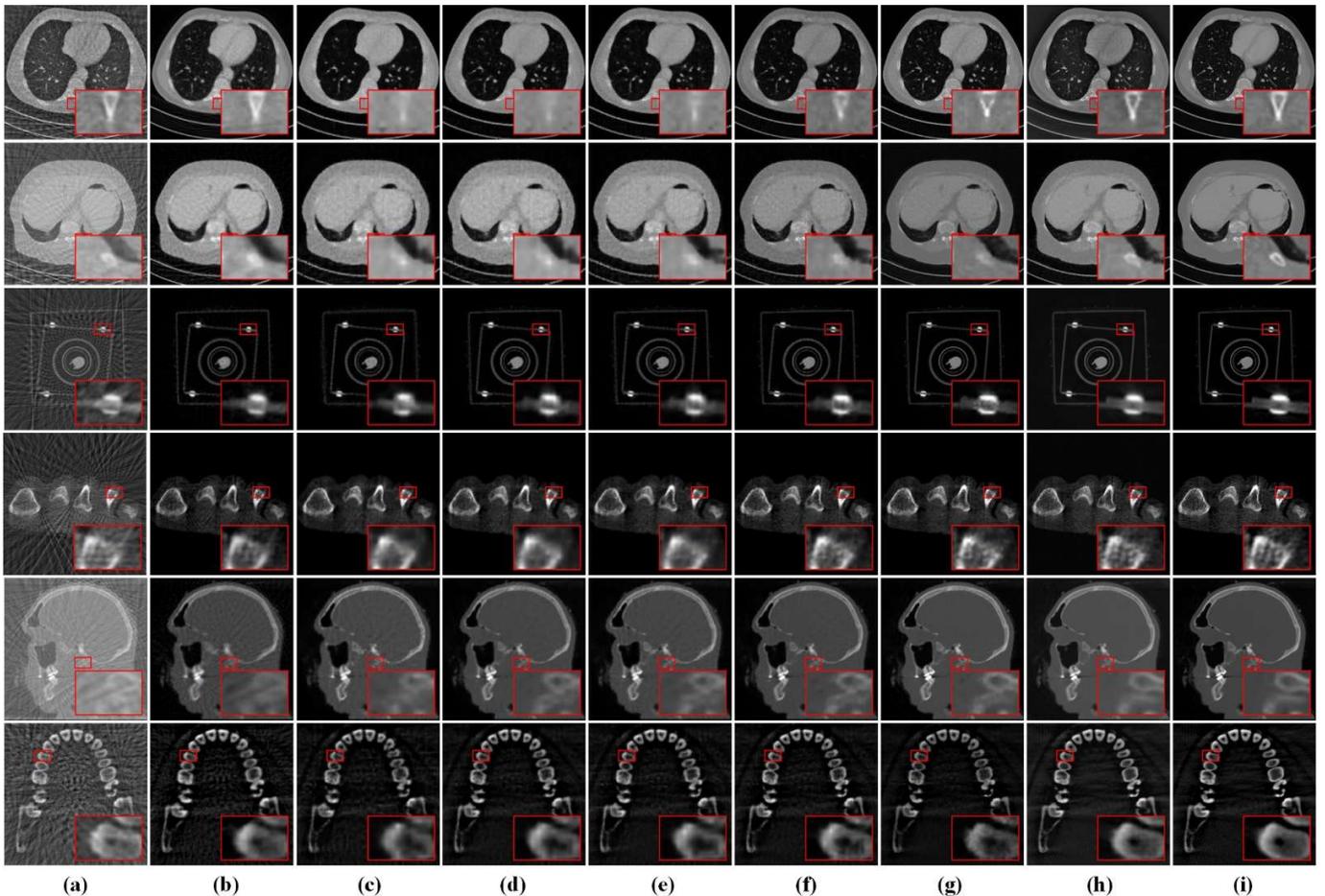

Fig. 5. Qualitative CT reconstruction comparison under the 50-view setting. Representative slices reconstructed by eight methods are shown. The vertical axis lists five datasets: AAPM (L067 and L096), Box, Head, Jaw, and Foot. The horizontal axis, from (a) to (i), corresponds to FDK, SART, InTomo, NeRF, NeAT, NAF, SAX-NeRF, Ours, and the ground truths.

*c) Training and Implementation Details:* For both the public and experimental data validation, weighted reconstruction was performed under two distinct protocols to evaluate robustness: an ultra-sparse configuration with 23 views and a sparse configuration with 50 views. The NAF prior was optimized with 1500 gradient updates. In the sinogram domain pathway, the Sino-RD diffusion model was trained using sinogram projections from 512 views over 1500 epochs. Similarly, in the DR pathway, the DR-RD model was trained using DR projections from 720 views, also for 1500 epochs. All experiments were executed on an NVIDIA A6000 GPU. Our implementation has been publicly released at: https://github.com/yqx7150/CSDN.

Table II
QUANTITATIVE CT RECONSTRUCTION RESULTS UNDER THE 23-VIEW SETTING. ENTRIES ARE PSNR/SSIM ON SIX DATASETS.

| Method/Dataset | L067 | L096 | Box | Foot | Head | Jaw |
| --- | --- | --- | --- | --- | --- | --- |
| FDK | 22.23/0.5157 | 24.59/0.5916 | 22.80/0.5472 | 19.73/0.3830 | 24.29/0.6291 | 22.81/0.4586 |
| SART | 27.11/0.7637 | 29.13/0.7336 | 31.30/0.9206 | 28.64/0.9052 | 27.88/0.8980 | 28.80/0.8235 |
| InTomo | 29.80/0.8951 | 31.54/0.9169 | 31.90/0.9290 | 29.77/0.9078 | 33.67/0.9559 | 30.45/0.8675 |
| NeRF | 27.25/0.8364 | 28.48/0.8455 | 30.81/0.9074 | 28.90/0.8993 | 29.65/0.8919 | 28.25/0.7963 |
| NeAT | 27.99/0.8465 | 30.08/0.8877 | 32.18/0.9371 | 29.78/0.9093 | 31.81/0.9276 | 29.91/0.8437 |
| NAF | 29.49/0.8871 | 31.43/0.9141 | 33.22/0.9515 | 29.63/0.9060 | 31.96/0.9281 | 30.42/0.8573 |
| SAX-NeRF | 30.50/0.8981 | 33.43/0.9389 | 33.40/0.9534 | 29.94/0.9117 | 33.45/0.9560 | 30.68/0.8723 |
| **Ours** | **32.44/0.9367** | **35.14/0.9626** | **35.46/0.9768** | **30.95/0.9289** | **35.17/0.9684** | **32.25/0.9071** |

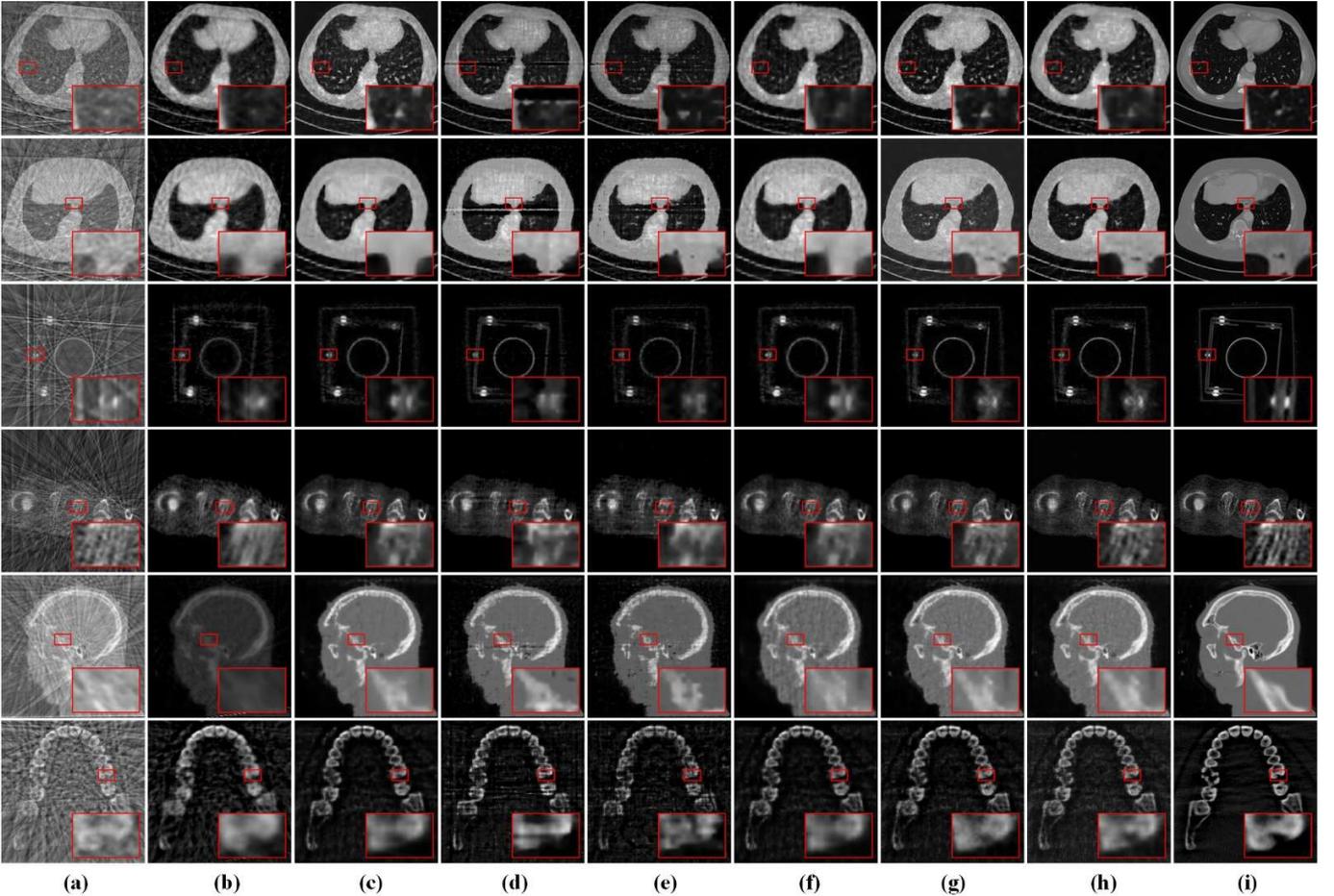

Fig. 6. Qualitative CT reconstruction comparison under the 23-view setting. Representative slices reconstructed by eight methods are shown. The vertical axis lists five datasets: AAPM (L067 and L096), Box, Head, Jaw, and Foot. The horizontal axis, from (a) to (i), corresponds to FDK, SART, InTomo, NeRF, NeAT, NAF, SAX-NeRF, Ours, and the ground truths.

### B. Evaluation Metrics

This study employs the Structural Similarity Index (SSIM) and the Peak Signal-to-Noise Ratio (PSNR) as quantitative evaluation metrics. To assess the optimization performance of 2D DR projections, SSIM and PSNR are calculated for each individual image slice and then averaged.

For evaluating the reconstruction quality of 3D CT, both metrics are computed directly on the full volumetric data SSIM measures image quality from the perspective of structural similarity, while PSNR reflects signal fidelity based on the mean squared error. Higher values for both metrics indicate that the reconstructed results are closer to the reference data.

### C. Comparative Experiments

This study conducted 3D CT reconstruction experiments under 50-view and 23-view settings, systematically comparing the performance differences between the proposed method and various mainstream reconstruction algorithms. Both quantitative and qualitative results demonstrate that our method achieves high-quality reconstruction from projection data of varying sparsity levels.

For the 50-view reconstruction task, we compared traditional algorithms such as FDK and SART with several advanced learning-based methods, including InTomo, NeRF, NeAT, NAF, and SAX-NeRF. As shown in Table I, our method achieves superior reconstruction quality across all datasets, with PSNR improvements of 2.51 dB and 1.44 dB on the AAPM L067 and L096 datasets, respectively. The PSNR gains on the box, foot, head, and jaw datasets are 3.38 dB, 1.19 dB, 1.11 dB, and 3.30 dB, respectively. Visual comparisons in Fig. 5 further confirm the distinct advantages of our method in noise suppression and structural integrity preservation.

For the more challenging 23-view super-sparse reconstruction task, the performance comparison of all methods is summarized in Table II. Experimental data indicate that our method maintains its leading performance even with extremely limited projection data, achieving PSNR improvements of 1.94 dB and 1.71 dB on the AAPM L067 and L096 datasets, respectively. The PSNR gains on the box, foot, head, and jaw datasets reach 2.06 dB, 1.01 dB, 1.72 dB, and 1.57 dB, respectively. Fig. 6 provides visual evidence of significantly improved noise suppression and superior recovery of structural details under ultra-sparse sampling conditions. Even with ultra-sparse-view sampling with only 23 views, streak artifacts and other sparse-view artifacts are greatly reduced, and the anatomy is faithfully recovered.

The result obtained with our CSDN method is shown in Fig. 7. Even under this sparse 50-view condition, streak artifacts and structural distortions are substantially suppressed. The 3D volume rendering and the representative 2D slices demonstrate that our method consistently recovers the intricate exoskeletal morphology of the specimen, including fine textures and contours, with high fidelity. Crucially, the sharp interface between the biological tissue and the surrounding glass tube is also clearly preserved, indicating the method's capability to maintain high-contrast boundaries without introducing blurring or merging artifacts. This successful reconstruction from real, limited data underscores the effectiveness and generalizability of the CSDN framework's dual-path continuity refinement strategy beyond simulated datasets.

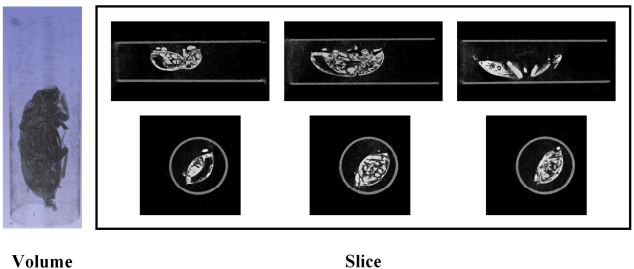

Fig. 7. CT reconstructions of a Lucanidae specimen placed in a glass tube under the 50-views setting. The left panel shows a 3D volume rendering demonstrating the recovered exoskeletal structure and the surrounding glass tube, and the right panel shows several representative axial and coronal slices reconstructed using the proposed CSDN method.

### D. Ablation Study

*a) Refine Model:* In this study, experiments were conducted under two configurations: 50 views and 23 views. Fig. 8 presents the distribution characteristics of the projection data before and after optimization under the 50-view condition using violin plots. Specifically, subfigures (a) and (b) illustrate the optimization effects on DR and sinogram, respectively. The results indicate that Sino-RD and DR-RD can effectively improve the distribution of the projection data.

*b) Dual-domain Weight:* To validate the effectiveness of the dual-path collaborative reconstruction strategy proposed in this paper, we designed three comparative experimental configurations: sinogram-path only, DR-path only, and the dual-path fusion scheme with weight $\omega$. All experiments were conducted under consistent 23-view and 50-view sampling conditions.

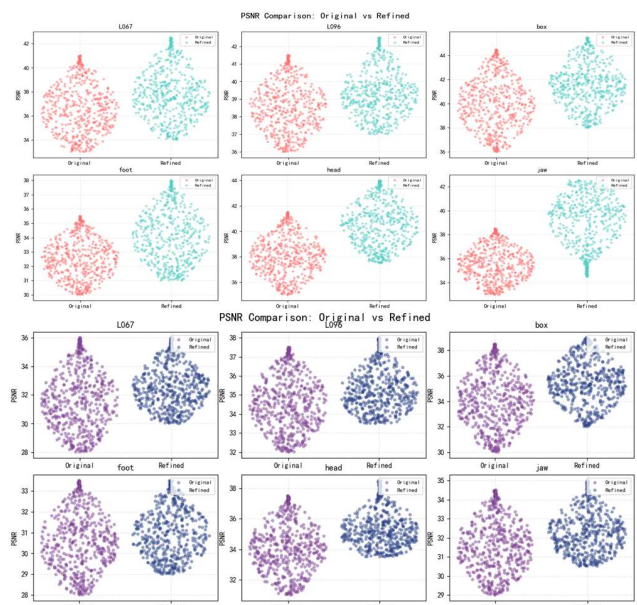

Fig. 8. The violin plots in this figure show the distribution characteristics of the projection data before and after optimization, under the 50-view condition. Subfigures (a) and (b) illustrate the effects of the Sino-RD and DR-RD optimization on the sinogram and DR projection, respectively.

In the sinogram-only setting, we employed a sinogram diffusion model to refine the initial sinogram, which was then reconstructed into the volume $V_{\text{sin}}$ via the FDK algorithm. In the DR-path only setting, a DR diffusion model was used to refine the initial DR projections, which were then reconstructed using the FDK algorithm to yield the volume data $V_{\text{dr}}$. In the dual-path fusion setting, the outputs of both paths were integrated according to the fusion mechanism proposed in this paper, with the weight set to $\omega = 0.8$.

We conducted a comprehensive visual analysis of the reconstruction performance using radar charts, and the quantitative results are summarized in Fig. 9. Under the 50-view sampling condition, the dual-path fusion method outperformed the single-path configurations across all datasets, achieving an average improvement in PSNR of over 0.43 dB and an increase in SSIM. Under the ultra-sparse 23-view condition, the dual-path approach still maintained a substantial advantage, with its average PSNR exceeding that of single-path methods by approximately 0.368 dB, demonstrating the effectiveness of this strategy even under extremely sparse sampling conditions.

### E. Parameter Experiment

*a ) Novel View Generation:* This study investigates the impact of the number of synthesized views $M$ on the reconstruction quality of NAF. Experiments were conducted under two uniformly sparse sampling configurations, namely 23-view and 50-view setups, and the performance was evaluated on a comprehensive test set. The results are illustrated in Fig. 10.

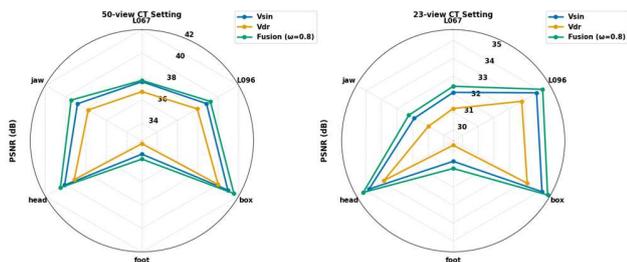

Fig. 9. The radar charts in this figure compare the PSNR performance of three reconstruction configurations: sinogram-path only, DR-path only, and dual-path fusion, under 23-view and 50-view sampling conditions.

In the experiments, we tested $M$ as 240, 480, 540, 640, 720, 800, 880, 960, 1200 and used the PSNR as the evaluation metric for image quality. The results show that under different sparse baselines, the reconstruction quality exhibits an initial increase followed by a decrease as the number of synthesized views grows. Taking the 50-view input condition as an example, when the number of synthesized views increased from 240 to 720, the PSNR improved by approximately 0.84 dB on average. This confirms that supplementing angular continuity through view synthesis positively contributes to reconstruction quality.

While the number of synthesized views $M$ exceeds 720, the PSNR begins to decline. This stems from the inherent approximation error of the NAF model based on sparse-view priors: moderate view synthesis can effectively enhance angular sampling density, but excessive synthesis leads to systematic replication and amplification of this approximation error across the angular domain. Consequently, significant artifacts and noise accumulate during the subsequent FDK reconstruction process, ultimately degrading reconstruction quality. All experiments were conducted under the full-sampling baseline of 720 view.

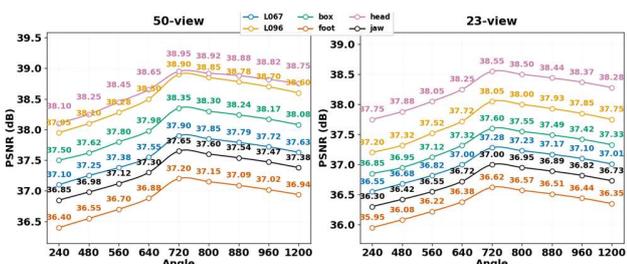

Fig. 10. The figure compares the PSNR performance of the NAF reconstruction model under 50-view and 23-view conditions. The results illustrate how reconstruction quality varies with $M$.

*b ) Weighted Fusion Coefficient Analysis:* This study systematically evaluates the impact of the weight coefficient $\omega$ in the dual-path fusion mechanism on reconstruction quality, examining values from 0.1 to 0.9 in increments of 0.1. The experiments were conducted across all six datasets under two sparse-view sampling scenarios: 23 views and 50 views, to ensure the generalizability of the conclusions.

Results demonstrate, as illustrated in Table III and Table IV, that as $\omega$ increases from 0.1 to 0.8, the PSNR shows a monotonic improvement. This trend underscores the critical role of the angular continuity prior provided by the sinogram pathway in suppressing artifacts induced by sparse sampling. This decline suggests that over-reliance on the sinogram pathway may compromise the inter-slice structural details preserved by the DR pathway, thereby validating the complementary design of the two paths. Notably, under the more challenging 23-view sampling condition, the optimal weight remains consistently around $\omega = 0.8$. This indicates that the proposed method effectively balances angular continuity with inter-slice consistency, even under conditions of severe data truncation, and exhibits strong robustness.

TABLE III
PSNR across datasets versus the fusion weight ω under 50-view sampling.

| DataSet/ω | 0.2 | 0.4 | 0.6 | 0.7 | **0.8** | 0.9 |
|---|---|---|---|---|---|---|
| L067 | 36.85 | 37.02 | 37.24 | 37.38 | **37.66** | 37.58 |
| L096 | 38.10 | 38.29 | 38.55 | 38.72 | **39.23** | 38.88 |
| Box  | 40.15 | 40.35 | 40.64 | 40.84 | **41.55** | 40.98 |
| Foot | 32.90 | 33.07 | 33.31 | 33.46 | **34.07** | 33.70 |
| Head | 39.30 | 39.50 | 39.78 | 39.97 | **40.51** | 40.13 |
| Jaw  | 37.92 | 38.11 | 38.37 | 38.54 | **39.45** | 38.88 |

TABLE IV
PSNR across datasets versus the fusion weight ω under 23-view sampling.

| DataSet/ω | 0.2 | 0.4 | 0.6 | 0.7 | **0.8** | 0.9 |
|---|---|---|---|---|---|---|
| L067 | 31.35 | 31.52 | 31.76 | 31.91 | **32.44** | 32.15 |
| L096 | 33.95 | 34.15 | 34.43 | 34.64 | **35.14** | 34.80 |
| Box  | 34.30 | 34.50 | 34.78 | 34.97 | **35.46** | 35.11 |
| Foot | 29.80 | 29.97 | 30.21 | 30.36 | **30.95** | 30.60 |
| Head | 34.00 | 34.20 | 34.48 | 34.66 | **35.17** | 34.83 |
| Jaw  | 31.15 | 31.32 | 31.56 | 31.71 | **32.25** | 31.95 |

## V. CONCLUSION AND DISCUSSIONS

This study addresses the challenges of angular discontinuity and inter-slice inconsistency in ultra-sparse-view sampling cone-beam CT reconstruction by proposing a neural prior-guided dual-path synergistic diffusion reconstruction framework named CSDN. The method first leverages a neural attenuation field to learn continuous 3D scene priors from extremely sparse projections, generating dense projection data. Subsequently, a dual-path diffusion module is employed to enhance the angular continuity and inter-slice consistency of the projection data separately. Finally, a dual-path weighted fusion mechanism integrates the two reconstruction pathways to achieve high-quality 3D reconstruction. Experimental results demonstrate that the proposed method significantly suppresses streak artifacts and restores structural details under extremely sparse sampling conditions such as 23-view and 50-view acquisitions, outperforming existing state-of-the-art methods in multiple quantitative metrics and visual quality. This study provides an effective approach for sparse-sampling CBCT reconstruction that integrates neural scene priors with dual-domain diffusion enhancement, contributing positively to the advancement of low-dose, high-precision three-dimensional imaging.